\documentclass{article}

\usepackage[preprint,nonatbib]{confstyle}

\usepackage[utf8]{inputenc} 
\usepackage[T1]{fontenc}    
\usepackage{hyperref}       
\usepackage{url}            
\usepackage{booktabs}       
\usepackage{amsfonts}       
\usepackage{nicefrac}       
\usepackage{microtype}      
\usepackage{xcolor}         
\usepackage{graphicx}  
\usepackage{setspace}
\usepackage{multirow}
\usepackage{hyperref}

\usepackage{mathptmx} 
\usepackage{times} 
\usepackage{amssymb}  

\usepackage{listings} 

\usepackage{algorithm}  
\usepackage{algorithmic}
\usepackage{amsmath}
\newtheorem{assumption}{Assumption}
\newtheorem{remark}{Remark}

\usepackage{balance}
\usepackage{adjustbox}
\usepackage{tabularx}

\def\sumtone{\sum_{t=1}^T}

\def\pr{\mathbb{P}}

\def\var{\mathrm{var}}
\def\API{neural signal}

\def\T{\mathrm{T}}

\newtheorem{problem}{Problem}

\title{Body Discovery of Embodied AI}

\author{Zhe Sun$^{1\ast}$, Pengfei Tian$^{2\ast}$, Xiaozhu Hu$^{3}$, Xiaoyu Zhao$^{4}$, Huiying Li$^{1}$, Zhenliang Zhang$^{1\dagger}$
\thanks{${\ast}$ Equal contributors}
\thanks{${\dagger}$ Corresponding author.  {\tt\scriptsize Email: zlzhang@bigai.ai}}
\thanks{$^{1}$ State Key Laboratory of General Artificial Intelligence, Beijing Institute for General Artificial Intelligence (BIGAI), Beijing, China. $^{2}$ Tsinghua University, Beijing, China. $^{3}$ Hong Kong University of Science and Technology, Hong Kong, China. $^{4}$ University of Rochester, New York, USA. }
}

\makeatletter
\def\thanks#1{\protected@xdef\@thanks{\@thanks
        \protect\footnotetext{#1}}}
\makeatother

\begin{document}

\maketitle

\begin{abstract}
In the pursuit of realizing artificial general intelligence (AGI), the importance of embodied artificial intelligence (AI) becomes increasingly apparent. Following this trend, research integrating robots with AGI has become prominent. As various kinds of embodiments have been designed, adaptability to diverse embodiments will become important to AGI. We introduce a new challenge, termed \textbf{``Body Discovery of Embodied AI''}, focusing on tasks of recognizing embodiments and summarizing neural signal functionality. The challenge encompasses the precise definition of an AI body and the intricate task of identifying embodiments in dynamic environments, where conventional approaches often prove inadequate. To address these challenges, we apply causal inference method and evaluate it by developing a simulator tailored for testing algorithms with virtual environments. Finally, we validate the efficacy of our algorithms through empirical testing, demonstrating their robust performance in various scenarios based on virtual environments.
\end{abstract}

\section{Introduction}
\label{sec:intro}

Intelligence is rooted in the interaction between embodied agents and the surrounding environments~\cite{peng2024tong}. 
As the development of artificial general intelligence (AGI) \cite{goertzel2007artificial, goertzel2014artificial, pei2019towards}, utilizing embodied agents is considered a plausible path to achieving AGI \cite{baum2017survey, duan2022survey,zhang2024emergence,zhang2019symmetrical}. AGI is expected to adapt to a diverse range of physical embodiments \cite{cully2015robots,shah2021soft,yu2008morpho}, breaking the restriction of their initial design. 
For instance, in smart homes \cite{alaa2017review,sun2023neighbor}, different households have varying configurations and contain different physical units, and a robust AGI is supposed to function effectively in all scenarios.

Adapting to different physical embodiments manifests the intelligent features of embodied AI \cite{gupta2021embodied}. To achieve this capability, two points should be considered: i) understanding the composition and boundaries of these embodiments, and ii) comprehending their functionalities. The body serves as the vessel for capabilities, and by recognizing the intricacies of one's physical embodiments, we can better maintain and utilize our abilities \cite{pfeifer2006body}. Furthermore, understanding the functionalities of these diverse physical embodiments \cite{lee2006physically} is essential for effectively handling a wide array of tasks.

In the fields of biology and human cognition, numerous researchers have devoted their efforts to understanding self-recognition and the connection between an organism and its environment \cite{cutting1982two,fodor1985precis,anderson2003embodied}. Several experiments, such as ``Mirror Test''~\cite{gallup2002mirror}, have been conducted to gauge capacity for self-cognition. In contrast, within the realm of robotics, there has been a growing emphasis on integrating AI algorithms with hardware to expand the spectrum of tasks AI systems can undertake \cite{driess2023palm,vemprala2023chatgpt,rajan2017towards}. However, a crucial yet relatively unexplored domain is the ability of AI systems to autonomously recognize their bodies (i.e., the physical embodiments), despite the significance of this research field being revealed. 
Although some researchers have attempted to address this topic, their focus has been limited to humanoid robots in relatively fixed settings \cite{saegusa2010self}, which cannot cover the complicated scenarios of embodied AI.

To the best of our knowledge, our proposal marks the pioneering attempt to autonomously discover and evaluate an AI system's body (both single and multiple agents) in general dynamic environments without the dependence on developers’ manual specifications. This is also beyond the research scope of general robot morphology recognition~\cite{diaz2023machine}, which is focused on self-discovery of only the single robot body morphology. The consideration of multiple agents obviously increase the difficulty because many kinds of environmental objects will interference the AI system's self detection.

We introduce a new problem named ``\textbf{Body Discovery of Embodied AI}''.
This problem comprises two primary tasks: the first revolves around recognizing the embodiments, and the second involves summarizing the functionality of each neural signal (namely the signals emitted by AI to control its body) that is linked to the embodiments.
For clarity of terminology in this paper, ``AGI'' represents the next milestone of ``AI'', and the ``Embodied AI'' equipped with the ability to recognize its own ``body'' (i.e., embodiment) exhibits a promising path to achieving AGI.

\begin{figure}
    \centering
    \includegraphics[width=1\linewidth]{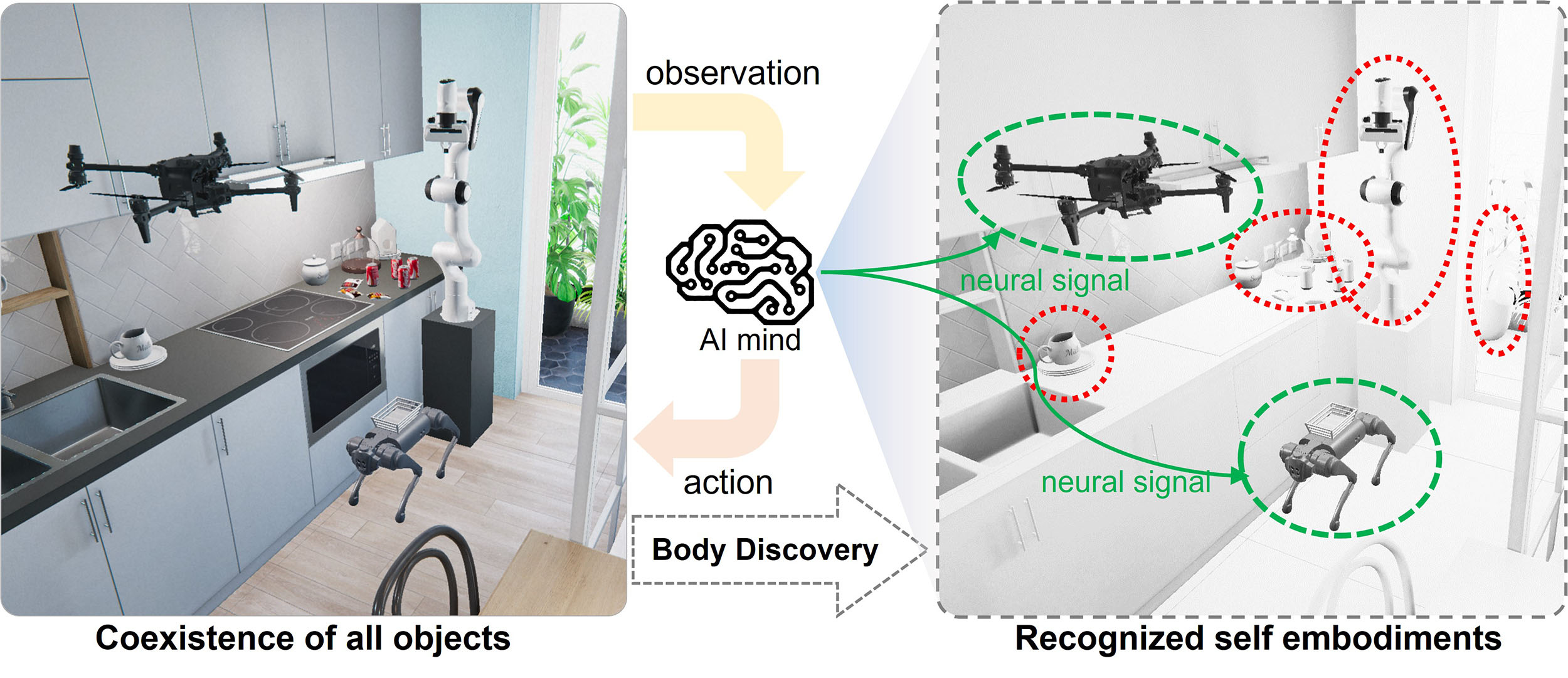}
    \vspace{-12pt}
    \caption{The ability to recognize embodiments is meaningful to embodied AI agents. With such ability, agents can adapt to various sets of embodiments through self-experimentation when being put into new scenarios.}
\label{fig:teaser}
\end{figure}

The problem at hand presents numerous hurdles.
First, the initial obstacle is to provide a cogent definition of an AI body and to outline the specific problem it poses precisely. When an AI system suffers from body damage or has to transfer into another purely new set of hardware with different structures, it cannot directly adapt to new configurations of bodies, which prohibits the normal functions of AI.
Second, another challenge pertains to identifying embodiments in a dynamic environment. Conventional supervised learning approaches falter when dealing with unlabeled data, and reinforcement learning struggles due to the diversity of body types in different environments. More specifically, the desired AGI should possess the capability to adapt to a newly transferred body, which may entail different functionalities and diverse environments. Furthermore, environmental noise and other AI agents' unpredictable behaviors compound the difficulty of distinguishing the true body of a specific AI agent.

In this paper, we define the problem called ``Body Discovery of Embodied AI'' and apply the causal inference method to the problem, as shown in Figure~\ref{fig:teaser}. In summary, our contributions are three-fold:
\begin{itemize}
    \item We delve into the embodiment of artificial intelligence and formulate the body discovery challenge called ``Body Discovery of Embodied AI''.
    \item 
    We introduce a causal inference framework to rewrite the problem and apply method enabling the detection of the body and the summarization of its functions.
    \item Through experiments conducted in a simulated environment, we showcase the efficacy of the developed algorithm for embodied AI encompassing single and multiple agents, highlighting its capabilities in solving the proposed problem.
\end{itemize}

\section{Problem Formulation}

In body functionalism, the body is defined as the vehicle for executing functions \cite{kiverstein2012meaning,wilson2002six,anderson2003embodied}. 
Human minds \cite{minsky1988society} generate neural signals and drive the body to take a corresponding physical response. Analogously, AI minds create neural signals to drive the embodiments. The AI mind here refers to the computational module of an AI system, which processes the perceived information and generate control signals to plan agent behaviors.

The problem of ``Body Discovery of Embodied AI'' is focused on i) recognizing the embodiments that are driven by the AI mind in a given environment, and ii) figuring out the functions that are associated with the embodiments. 
Specially speaking, when adapting to a new physical configuration, AI can generate some neural signals to connect to the new physical embodiments (or say the AI's body), and we denote the number of neural signals as $Q$. In this situation, AI should comprehend both its new neural architecture (i.e., the relationship between neural signals and bodies) and the functions associated with $Q$ neural signals (i.e., the effects that the embodiments can produce).

\begin{figure}
    \centering
    \includegraphics[width=\linewidth]{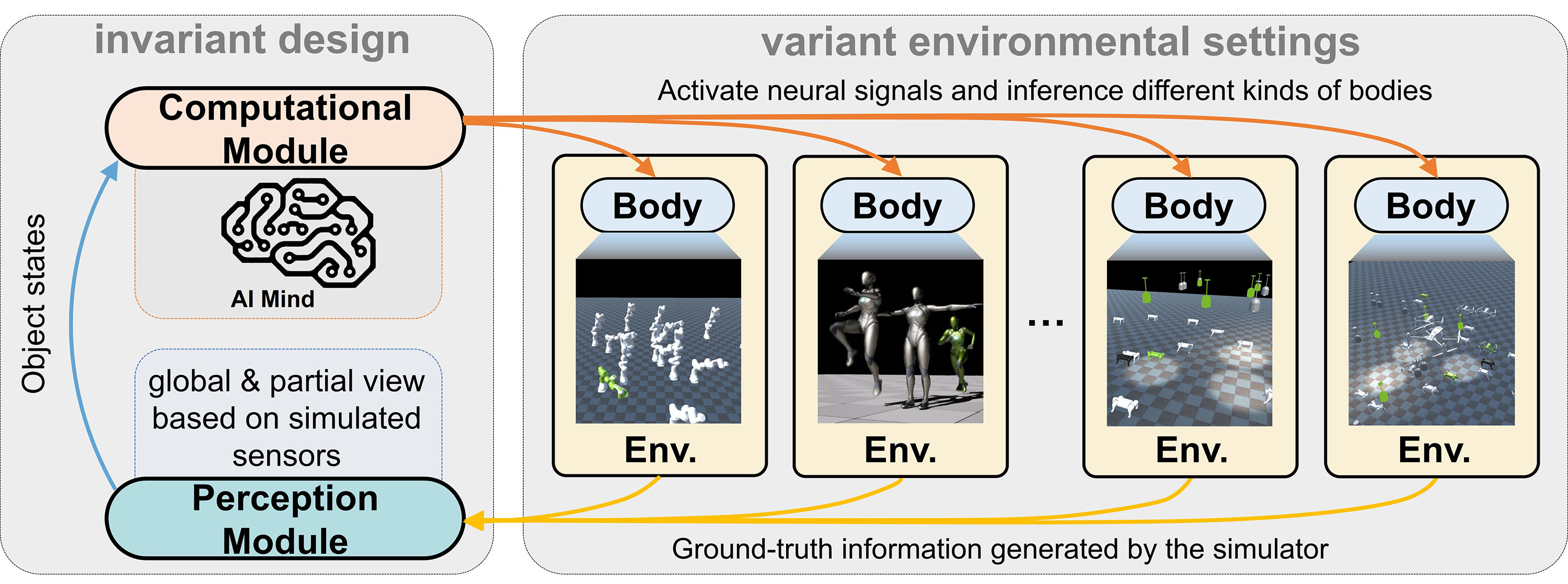}
    \vspace{-12pt}
    \caption{Framework of the body discovery challenge. The design of the agent is assumed to be invariant across different tasks. Composed of a computational module (i.e., AI mind) and a perception module, the agent is compatible with various embodiments and can be implemented in different world settings. The category of embodiments can be different or even time-varying. }
    \label{fig:framework}
\end{figure}

To address this problem, AI needs to perceive and understand its surroundings. As shown in Figure~\ref{fig:framework}, AI is supposed to be able to detect its own body structure (both single and multiple agents) in variant environmental settings. Ideally, it would have access to global observations, but in practice, it often has only a partial view. In our research, the partial view is simulated by involving flaws in global observations, which can reflect the practical observation to a large extent.

\subsection{\textbf{Problem Description}}
In this context, we assume all objects $O_1, \ldots, O_N$ in the space can be observed. Among them, some are parts of the body, and we denote them as body composition set $\mathcal{B}$. Each object is characterized by $K$ features, with the $k$-th feature of object $n$ denoted as $O_{nk}, n=1, \ldots, N, k=1, \ldots, K$. These features may encompass diverse information types, such as location and temperature.
Our focus is twofold, as articulated through the following questions:

\begin{problem}[Body Discovery of Embodied AI]{}\label{prob}
\ 
\quad
\begin{itemize}
    \item Among $O_1$, $O_2$,\ldots, $O_N$, what are the body composition $\mathcal B$ of AI in this environment?
    \item How to describe the effect of each neural signal when AI is controlling its body?
\end{itemize}
\end{problem}

To simplify the problem, we adopt discrete space to describe it.
We discretize the time period, assuming there are $T+1$ stages. The value of feature $k$ for object $n$ at the $t$-th stage is denoted as $S_{nkt}$, and the features at the $t$-th stage are collectively represented as $\mathcal{S}_t = [S_{nkt}]_{n,k}$. The initial stage, denoted as $\mathcal{S}_0$, marks the starting point.
For stages $t \geq 1$, artificial intelligence faces $Q+1$ choices: opting to generate no \API, denoted as Action $0$, or generating one of $Q$ distinct neural signals, denoted as Actions $1$ through $Q$. We exclude choosing multiple neural signals simultaneously, as each instance is considered a distinct \API. With only $T$ stages for action, the action array is denoted as $D = (D_1, \ldots, D_T)^\T \in \{0, 1, \ldots, Q\}^T$. For convenience, for any index set $\mathcal{I}$, $D_\mathcal{I}$ denotes the subvector of $D$ with dimensions indexed only in $\mathcal{I}$. Specifically, $t_1:t_2$ represents the index set $\{t_1, t_1+1, \ldots, t_2\}$ for any $t_1 < t_2$.

At last, the objective is to formulate a policy that determines $D$, and based on both $D$ and the observed space $\{\mathcal{S}_t\}_{t=0}^T$, infers the identity of AI body among these objects and summarizes the neural signal function.

\subsection{\textbf{Problem Solving}}
It is important to acknowledge the inherent complexity of Problem \ref{prob}. 
Owing to the intricacies of the real environment, the impact of each neural signal is entangled with noise, and features may encounter interference. Additionally, data inference challenges such as selection bias and confounding factors are prevalent. Introducing a causal inference framework becomes imperative to articulate the problem clearly.
 
Here we adopt the potential outcome framework \cite{Neyman:1923,Rubin:1974}. Let $S_{nkt}(D_{1:t})$ represent the $k$-th feature value of object $n$ at the $t$-th stage when the intelligence utilizes policy $D_{1:t}$. The observations collected at stage $t$ under policy $D_{1:t}$ are denoted as $\mathcal{S}_t(D_{1:t})=[S_{nkt}(D_{1:t})]_{n,k}$. Then the $t$-th stage change is $\Delta_t(D_{1:t})=\mathcal S_t(D_{1:t})-\mathcal S_{t-1}(D_{1:{t-1}})$. To handle the complexity, we introduce the Stable Stage Treatment Value Assumption (SSTVA) as follows.

\begin{assumption}[SSTVA]\label{ass:sutva}
For stage $t=1,\ldots,T$, we assume that $\{\Delta_t\}_{t=1}^T$ satisfies following two conditions:  
\begin{itemize}
\item \textbf{(No Interference)}
$\Delta_t$ does not rely on other stages’ actions.
\item \textbf{(Consistency)} The action levels are well-defined and do not have ambiguous meanings for outcomes of interest.
\end{itemize}
\end{assumption}

\begin{remark}
Assumption \ref{ass:sutva} parallels the foundational causal inference assumption, the Stable Unit Treatment Value Assumption \cite{Rubin:1974,imbens2015causal}, but extends the concept of stability from units to stages. This modification is particularly relevant for robotics, where the notion of ``No Interference'' is reasonable, with the assumption that each neural signal persists for a single stage. Most actions, especially movements, conform to the no interference assumption.
\end{remark}

\begin{remark}
Under Assumption \ref{ass:sutva}, ``Consistency'' ensures the stability of the neural signal function, eliminating ambiguity and serving as a natural assumption, so we can express $\Delta_t = \Delta_t(D_{1:t}) = \Delta_t(D_t)$.
For instance, a cluster of drones flies in the sky, where the wind direction remains stable within each stage, and the control signals for each drone consistently maintain stability, remaining unchanged across stages.
\end{remark}

See Table \ref{tab:notation} for previously mentioned notations.

\section{Algorithm}
Our goal is to develop an algorithm that can handle the discovery of bodies, even in a new environment without prior knowledge, data, or models to reference. Thus, traditional supervised learning that relies heavily on labeled data cannot be used. Additionally, the operating mode and action patterns are unknown, making it impossible to apply reinforcement learning (RL) directly. We introduce the causal inference approach considering each stage as a ``unit''. Details are shown in Algorithm~\ref{alg:BD}.

\begin{table}[tb]
    \centering
    \caption{Body Discovery Framework Notation}
    \begin{tabular}{c|c}
    \toprule
    Notation & Meaning\\
    \midrule
        $Q$ & number of neural signals \\
        $N$ & total object number \\
        $K$ & total feature number\\
        $T$ & total stage number\\
        $O_n$ & Object $n$\\
        $S_{nkt}$ & the features $k$ of object $n$ at the $t$-th stage\\
        $\mathcal{S}_t$ & the features of all objects at the $t$-th stage\\
        $D_t$ & $t$-th stage neural signal action\\
        $n_q$ & total number of allocating action $q$ \\
        $\hat\xi_q^{(n,k)}(D)$ & difference-in-mean estimator\\
        $\Delta_t(D)$& $t$-th stage change  \\
    \bottomrule
    \end{tabular}
    \label{tab:notation}
\end{table}

\subsection{\textbf{Phase I: Randomized Experiment}}
The randomized experiment is widely recognized as the gold standard in causal inference, excelling in eliminating confounding variables and yielding precise estimators \cite{Neyman:1923}. We prioritize randomized experiments over decision richness from previous stages, choosing a more robust and causality-focused framework.

Choose $n_0, n_1, \ldots, n_Q \ge 1$, satisfying the condition $n_0 + \ldots + n_Q = T$. Denote our design allocation as $\mathbf{D} \in \{0,1, \ldots, Q\}^T$, representing a random vector. Any treatment $D$ is randomly selected, containing $n_0$ instances of Action 0, $n_1$ instances of Action $1$, ..., and $n_Q$ instances of Action $Q$. In other words,
$$\pr(\mathbf{D} = D) = \frac{n_0!n_1!\cdots n_Q!}{T!}.$$
\vspace{-4pt}

Denote $\Delta_t^{(n,k)}(q)$ as the value of Feature $k$ of Object $n$ under Action $q$ at Stage $t$ for $q=0,1,\ldots,Q$.
In the analysis phase, the difference between Action $q$'s and Action $0$'s stage changes (control group) reflects the average effect of Action $q$:
$$\hat\xi_q^{(n,k)}(D)=\frac{1}{n_q}\sumtone \Delta_t^{(n,k)} 1(D_t=q)-\frac{1}{n_0}\sumtone\Delta_t^{(n,k)} 1(D_t=0).$$
When ensuring the object belongs to the body, $\hat\xi_q^{(n,k)}(D)$ can summarize the effect of Action $q$ and is an unbiased estimator of the commonly considered causal target:
$$\xi_q^{(n,k)}(D)=\frac{1}{T}\sumtone \left(\Delta_t^{(n,k)}(q) -\Delta_t^{(n,k)}(0)\right).$$
The variance of the estimator $\hat\xi^{(n,k)}_q(D)$ equals \cite{imbens2015causal}:
$$\var\left(\hat \xi_q^{(n,k)}\right)=\frac{(S_q^{(n,k)})^2}{n_q}+\frac{(S_0^{(n,k)})^2}{n_0}-\frac{(S_{\tau q}^{(n,k)})^2}{n_q+n_0},$$
where $(S_{\tau q}^{(n,k)})^2=(T-1)^{-1}\sumtone\{(\Delta_t^{(n,k)}(q)-\bar \Delta^{(n,k)}(q))-(\Delta_t^{(n,k)}(0)-\bar \Delta^{(n,k)}(0))\}^2$ and $(S_{q}^{(n,k)})^2=(T-1)^{-1}\sumtone\{\Delta_t^{(n,k)}(q)-\bar \Delta^{(n,k)}(q)\}^2$. 
Because $\Delta_t^{(n,k)}(q)$ and $\Delta_t^{(n,k)}(0)$ cannot be observed simultaneously at each stage, $(S_{\tau q}^{(n,k)})^2$ cannot be identified and the traditional confidence interval is not exact. We introduce the Fisher randomization test \cite{imbens2015causal,ding2024first} to conduct exact inference under a small-scale trial.

\subsection{\textbf{Phase II: Fisher Randomization Test}}
We assess whether neural signal $q$ controls Feature $k$ of Object $n$ bt Fisher randomization test. Fisher randomization test operates under the sharp null hypothesis:
$$H_0: \Delta_t^{(n,k)}(q)=\Delta_t^{(n,k)}(0),\quad t=0,\ldots,T.$$
The alternative hypothesis posits the existence of a $t\in\{1,\ldots,T\}$ such that $\Delta_t^{(n,k)}(q)\neq\Delta_t^{(n,k)}(0)$. The Fisher randomization test utilizes the mentioned statistic $\hat\xi_q^{(n,k)}(D)$ dependent on the neural signal sequence $D$.

\begin{figure}[tb]
\setstretch{1}
\begin{algorithm}[H]
   \caption{Causality-Based Body Discovery}
   \label{alg:BD}
\begin{algorithmic}\small
   \STATE {\bfseries Input:} Confidence level $1-\alpha$, Monte Carlo parameter $M$, trial number $T$, neural signal number $Q$,
   number of stages $n_q$ ($q \in \{0,\ldots,Q\}$, caused by action $q$), 
   total object number $N$. 
   \STATE {\bfseries Output:} Predicted body, and the effect of each neural signal. 
    \STATE  {\bfseries Generate neural signals sequence:} Generate randomized sequence $D\in\{0,\ldots,Q\}^T$,    
    \STATE {\bfseries Record updated states:} $T+1$ Stages $\{\mathcal S_t=[S_{nkt}]_{n,k}\}_{t=0}^T$.
    \FOR{each $1\le q\le Q$, $1\le n\le N$ and $1\le k \le K$}
    \STATE Compute effect 
    $\hat \xi_q^{(n,k)}(D)=n_q^{-1}\sumtone \Delta_t 1(D_t=q)-n_0^{-1}\sumtone \Delta_t 1(D_t=0)$. 
    \FOR{$m=1$ to $M$}
    \STATE Generate $D^{(m)}$: Fix $D=(D_1,\ldots,D_T)^\T$'s dimension whose value equals $0$ or $q$, and permute $D$
    \STATE $\hat\xi_q^{(n,k)}(D^{(m)})=n_q^{-1}\sumtone \Delta_t 1(D_t^{(m)}=q)-n_0^{-1}\sumtone \Delta_t 1(D_t^{(m)}=0)$. Note, $\hat\xi_q^{(n,k)}(D)=(\hat\xi_q^{(n,k)})^{obs}$.
    \ENDFOR
    \STATE Compute p-value $p=\#\{m:|\hat\xi_q^{(n,k)}(D^{(m)})|\ge |(\hat\xi_q^{(n,k)})^{obs}|\}/M$.
    \IF{$p\le \alpha$ (or $\alpha/testing~number$\footnotemark[1])}
    \STATE Object $n$ belongs to the body.
    \ENDIF
    \ENDFOR 
\STATE {\bfseries return:} Predicted body list, and effect list $[\hat\xi_q^{(n,k)}]$.
\end{algorithmic}
\end{algorithm}
\vspace{-12pt}
\footnotesize{$^1$ We employ the Bonferroni correction method to address the common issue associated with multiple testing.} 
\end{figure}

Each stage can only apply one action, leading to the missingness of other potential outcomes. Fisher randomization test fills the missing data under the sharp null hypothesis based on our current observed sequence denoted as $D^{obs}$. We generate $\{\mathbf D^{(m)}\}_{m=1}^M$ based on $D^{obs}$, such that for $m=1,\ldots,M$, $\mathbf D^{(m)}$ only randomly permutes the positions of Action $0$ and $q$ while fixing other action positions in $D^{obs}$. For example, for sequence $D^{obs}=(0,1,2,1)^\T$, three kinds of permutations can be generated when studying Action $q=1$: $D^{(1)}=(0,1,2,1)$, $D^{(2)}=(1,0,2,1)$, $D^{(3)}=(1,1,2,0)$.

Subsequently, we compute the $p$-value indicating the significance whether neural signal $q$ controls Feature $k$ of Object $n$ by:
$$p(n,k,q)=\pr\left(|\hat\xi_q^{(n,k)}(\mathbf D^{(m)})|\ge |\hat\xi^{obs}|\right).$$
This $p$-value is exact and can be approximated by Monte Carlo as mentioned in Algorithm~\ref{alg:BD}. 

We set a commonly used critical value, such as $0.05$ or $0.01$. When the $p$-value is smaller than this threshold, we reject the null hypothesis and conclude that the neural signal controls the feature of the object.

In addition, when testing multiple hypotheses simultaneously, the original critical p-value may become unreliable. Therefore, we adopt the Bonferroni correction, replacing $\alpha$ with $\alpha/testing~number$ to address this issue.
As mentioned earlier, we obtain the summarized level function of neural signal $q$, $q=1,\ldots,Q$, and gain insights into how neural signal $q$ affects Feature $k$ of its body $n, n\in \mathcal B$ on average.

\begin{figure}[tb]
    \centering    
    \includegraphics[width=0.9\linewidth]{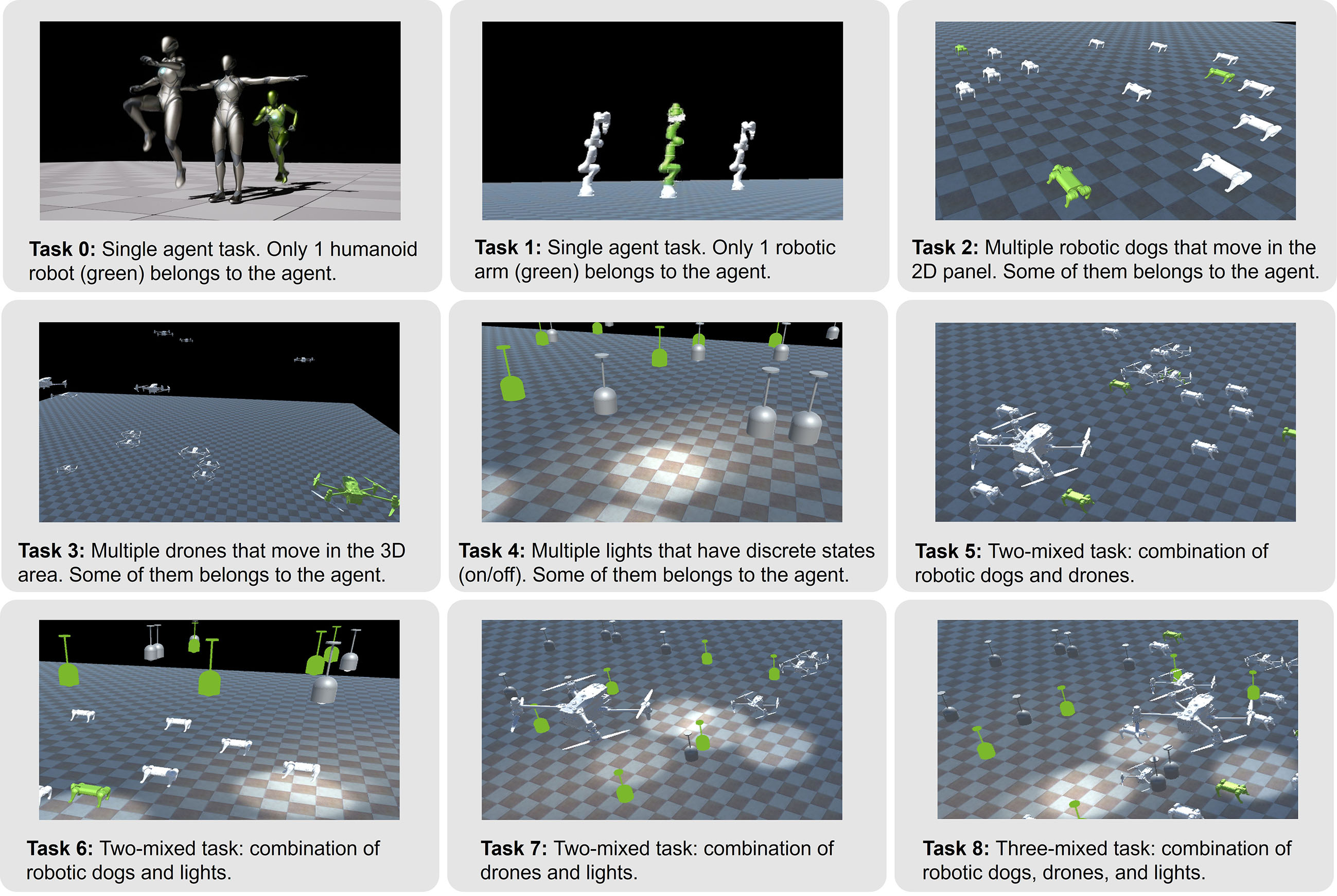}
    \caption{Simulated scenes for T0-T8. These tasks correspond to the 9 tasks in Table~\ref{tab:scenes}.}
    \label{fig:simulator_0_8}
\end{figure}

\begin{table}[t!]
  \caption{Categories of basic features included in T0-T8.}
  \centering
  \label{tab:scenes}
  \begin{tabular}{rl}
    \toprule
    Task & Category of feature\\
    \midrule
    \textit{(Single agent)} Humanoid robot (T0) & \textbf{Discrete state} of poses.  \\
    \textit{(Single agent)} Robotic arm (T1) & \textbf{Rotation} of joints.  \\
    Cluster of robotic dogs (T2)     & \textbf{2D position} of each entity.\\
    Cluster of drones (T3)      & \textbf{3D position} of each entity.\\
    Cluster of lights (T4)      & \textbf{Discrete state} of lights\footnotemark[1].\\
    Robotic dogs and drones (T5)      & \textbf{2D position} \& \textbf{3D position}.\\
    Robotic dogs and lights (T6)   & \textbf{2D position} \& \textbf{discrete state}.\\
    Drones and lights  (T7)      & \textbf{3D position} \& \textbf{discrete state}.\\
    Dogs, drones, and lights (T8) & \textbf{2D position} \& \textbf{3D position} \\
     ~ & \& \textbf{discrete state}. \\
    \bottomrule
    \end{tabular}
    \\
    \footnotesize{$^1$ The agent cannot turn on a light that is already on.} 
\end{table} 

\section{Experiments}

\subsection{Basic Experiments: Algorithm Analysis}

\subsubsection{\textbf{Experimental Goal}}
The experiments are designed to evaluate the method from the following perspectives:
\begin{enumerate}
    \item [i.] How do $Q$, $T$, and $N$ affect the performance?
    \item [ii.] How do the kind and intensity of noises affect the performance?
    \item [iii.] Performance among different task settings.
    \item [iv.] Can our method pass the mirror test?
\end{enumerate}

To evaluate the method, we consider 5 metrics commonly used in the machine learning field: the accuracy, the recall rate, the precision, the average precision, and the F1 score.

\subsubsection{\textbf{Experimental Setup}}
We design 4 typical categories of basic features: (1) rotation; (2) 2D position; (3) 3D position; (4) discrete state. 
Combinations of these basic features formulate 9 tasks (T0-T8). T0-T4 each includes one feature, while T5-T8 includes multiple features.
Table~\ref{tab:scenes} shows the details of each task and the features included.
Four kinds of noise are involved to simulate a relatively complex testing scenario. They are denoted as follows:
\begin{itemize}
    \item (N1) Environmental noise: the environment causes some object's states to change, e.g.: wind affects the position of the drones.
    \item (N2) Other agents: other agents in the environment could control some of the objects, either randomly or following specific patterns.
    \item (N3) Action failure: the execution of action fails.
    \item (N4) Sensing flaw: data from sensors has noise. 
\end{itemize}

\begin{table}[tb]
\setstretch{0.9}
\setlength\tabcolsep{1.5pt}
\caption{Performance of our methods and baselines across T0-T8.}
\vspace{-6pt}
\label{tab:baseline_compare}
\begin{center}
\resizebox*{0.7\linewidth}{!}{
\begin{tabular}{c|c|ccccc}
\toprule
Task & Method & Accuracy & Recall & Precision & Specificity & F1 Score \\
\midrule
\multirow{5}{*}{T0}
    & Baseline ($\alpha=0.05$) & 0.713 & 0.140 & 0.140 &\textbf{1.000} & 0.140 \\
    & Baseline ($\alpha=0.01$) & 0.680 & 0.040 & 0.040 & \textbf{1.000} & 0.040 \\
    & Ours ($p<0.05$) & \textbf{0.810} & \textbf{0.800} & \textbf{0.663} & 0.815 & \textbf{0.708} \\
    & Ours ($p<0.01$) & 0.803 & 0.720 & 0.618 & 0.845 & 0.652 \\
    & Ours (Bonferroni) & 0.777 & 0.560 & 0.515 & 0.885 & 0.530 \\
\midrule
\multirow{5}{*}{T1}
    & Baseline ($\alpha=0.05$) & 0.800 & 0.000 & 0.000 &\textbf{1.000} & N/A \\
    & Baseline ($\alpha=0.01$) & 0.800 & 0.000 & 0.000 & \textbf{1.000 }& N/A \\
    & Ours ($p<0.05$) & 0.916 & \textbf{0.982} & 0.257 & 0.914 & 0.407 \\
    & Ours ($p<0.01$) & \textbf{0.953} & 0.981 & 0.385 & 0.952 & 0.553 \\
    & Ours (Bonferroni) & 0.854 & 0.973 &\textbf{ 0.650} & 0.812 & \textbf{0.779} \\
\midrule
\multirow{5}{*}{T2} 
    & Baseline ($\alpha=0.05$) & 0.596 & \textbf{0.995} & 0.293 & 0.543 & 0.453 \\
    & Baseline ($\alpha=0.01$) & 0.725 & 0.984 & 0.416 & 0.694 & 0.585 \\
    & Ours ($p<0.05$) & 0.907 & 0.971 & 0.594 & 0.896 & 0.737 \\
    & Ours ($p<0.01$) & 0.974 & 0.960 & 0.874 & 0.976 & \textbf{0.914} \\
    & Ours (Bonferroni) & \textbf{0.984} & 0.882 & \textbf{0.946} & \textbf{0.999} & 0.913 \\
\midrule
\multirow{5}{*}{T3} 
    & Baseline ($\alpha=0.05$) & 0.446 & \textbf{0.998} & 0.259 & 0.393 & 0.411 \\
    & Baseline ($\alpha=0.01$) & 0.575 & 0.997 & 0.359 & 0.541 & 0.528 \\
    & Ours ($p<0.05$) & 0.891 & 0.997 & 0.528 & 0.878 & 0.690 \\
    & Ours ($p<0.01$) & 0.979 & 0.987 & 0.863 & 0.978 & 0.921 \\
    & Ours (Bonferroni) & \textbf{0.997} & 0.973 & \textbf{0.990} & \textbf{1.000} & \textbf{0.981} \\
\midrule
\multirow{5}{*}{T4} 
    & Baseline ($\alpha=0.05$) & 0.573 & \textbf{1.000} & 0.260 & 0.538 & 0.413 \\
    & Baseline ($\alpha=0.01$) & 0.670 & \textbf{1.000} & 0.328 & 0.646 & 0.494 \\
    & Ours ($p<0.05$) & 0.942 & \textbf{1.000} & 0.652 & 0.936 & 0.789 \\
    & Ours ($p<0.01$) & 0.986 & \textbf{1.000} & 0.880 & 0.985 & 0.936 \\
    & Ours (Bonferroni) & \textbf{0.999} & 0.997 & \textbf{0.995} & \textbf{0.999} & \textbf{0.996} \\
\midrule    
\multirow{5}{*}{T5} 
 & Baseline ($\alpha$=0.05) & 0.813 & 0.418 & 0.520 & 0.867 & 0.390 \\
 & Baseline ($\alpha$=0.01) & 0.815 & 0.386 & \textbf{0.530} & 0.874 & 0.371 \\
 & Ours ($p<0.05$) & 0.853 & \textbf{0.548} & 0.389 & 0.896 & 0.428 \\
 & Ours ($p<0.01$) & 0.891 & 0.510 & 0.500 & 0.945 & \textbf{0.474} \\
 & Ours (Bonferroni) & \textbf{0.898} & 0.454 & 0.519 & \textbf{0.960} & 0.459 \\
\midrule
\multirow{5}{*}{T6} 
 & Baseline ($\alpha$=0.05) & 0.796 & 0.439 & 0.450 & 0.848 & 0.375 \\
 & Baseline ($\alpha$=0.01) & 0.796 & 0.403 & 0.456 & 0.855 & 0.354 \\
 & Ours ($p<0.05$) & 0.811 & \textbf{0.560} & 0.314 & 0.846 & 0.382 \\
 & Ours ($p<0.01$) & 0.873 & 0.497 & 0.434 & 0.925 & \textbf{0.440} \\
 & Ours (Bonferroni) & \textbf{0.881} & 0.419 & \textbf{0.482} & \textbf{0.947} & 0.416 \\
\midrule
\multirow{5}{*}{T7} 
 & Baseline ($\alpha$=0.05) & 0.814 & 0.580 & 0.603 & 0.856 & 0.521 \\
 & Baseline ($\alpha$=0.01) & 0.812 & 0.556 & \textbf{0.607} & 0.860 & 0.504 \\
 & Ours ($p<0.05$) & 0.813 & \textbf{0.721} & 0.359 & 0.829 & 0.460 \\
 & Ours ($p<0.01$) & 0.882 & 0.703 & 0.508 & 0.912 & 0.563 \\
 & Ours (Bonferroni) & \textbf{0.901} & 0.699 & 0.588 & \textbf{0.935} & \textbf{0.603} \\
\midrule
\multirow{5}{*}{T8} 
 & Baseline ($\alpha$=0.05) & 0.803 & 0.501 & 0.582 & 0.858 & 0.470 \\
 & Baseline ($\alpha$=0.01) & 0.802 & 0.466 & \textbf{0.593} & 0.862 & 0.449 \\
 & Ours ($p<0.05$) & 0.826 & \textbf{0.645} & 0.386 & 0.857 & 0.467 \\
 & Ours ($p<0.01$) & 0.877 & 0.614 & 0.522 & 0.923 & 0.544 \\
 & Ours (Bonferroni) & \textbf{0.887} & 0.557 & 0.588 & \textbf{0.945} & \textbf{0.546} \\
\bottomrule
\end{tabular}
}\vspace{-12pt}
\end{center}
\end{table}

\subsubsection{\textbf{Procedures}}
We develop 9 simulated scenes for the 9 tasks, as shown in~Figure \ref{fig:simulator_0_8}. 
For each task, we test 10 rounds and calculate the average of each metric.
In every round, $N$ objects are generated inside the testing scene with random initial positions. Among them, some objects can be controlled by the agent through $Q$ neural signals. The effect of $q$ (controlled object ids and how $q$ changes their states) is randomly initialized and remains unchanged through its round.
$Q$ is given to the agent at the start of each round. The effect of each neural signal is unknown to the agent. The agent has $T$ times of attempts to find out which object belongs to it, as well as the effect of neural signals.

\subsubsection{\textbf{Evaluation}}

(i) \textbf{\textit{Statistics baseline.}} The baseline method directly calculates $\hat\xi_q^{(n,k)}$, the average effect of neural signal $q$ on feature $k$ of object $n$. For the set $\{\hat\xi_q^{(n,k)}\}$, we compute its mean $\bar\xi$ and the variance $\hat V_{\xi}$. According to \cite{LiDing2020}, these distributions follow a normal distribution asymptotically under the loose condition. Consequently, their sum also adheres to a normal distribution at the population level. Thus, we select critical values from the normal distribution's $1-\alpha/2$ quantile ($\alpha=0.01$ or $0.05$), denoted as $z_{1-\alpha/2}$. Then we identify body features that significantly deviate outside the confidence region $[\bar \xi-z_{1-\alpha/2} \hat V_{\xi},\bar \xi+z_{1-\alpha/2} \hat V_{\xi}]$.
(ii) \textbf{\textit{Parametric analysis.}}
We evaluate how the pre-defined settings and hyper-parameters affect performance: namely, the influence of neural signal number $Q$, candidate object number $N$, and total stage number $T$. We also test how the intensities of 4 kinds of noises (N1-N4) affect the performance. To make it clearer, we launched the analysis using T8, a relatively complex task with 3 kinds of features.

\begin{figure}[tb]
    \centering
    \includegraphics[width=\linewidth]{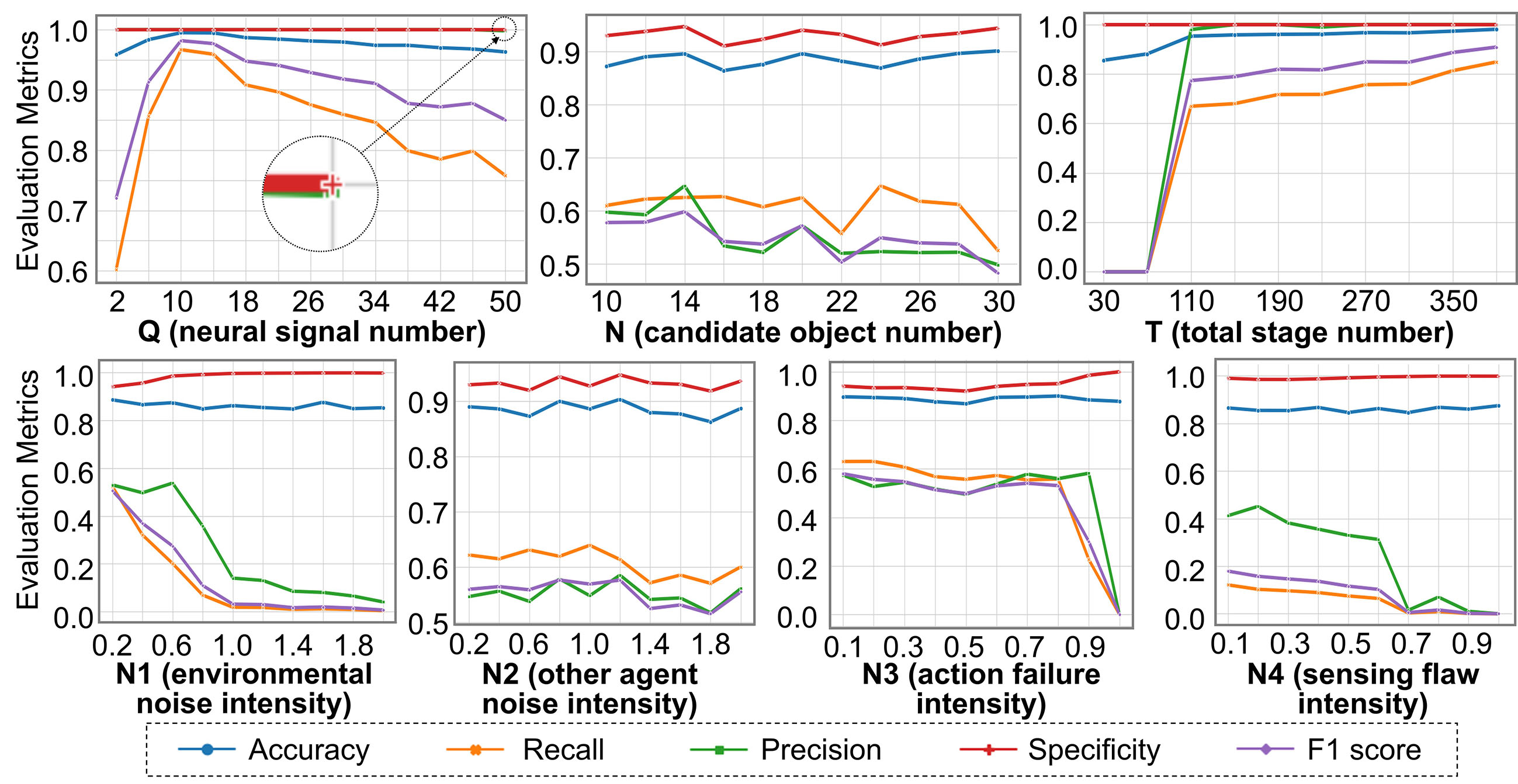}
    \vspace{-12pt}
    \caption{Results of parametric analysis. Performance changes along seven parameters on T8: Q (neural signal number), N (candidate object number), T (total stage number), N1 (environmental noise intensity), N2 (other agent noise intensity), N3 (action failure intensity), and N4 (sensing flaw intensity). The intensity of N1 and N2 is a calculated ratio. It indicates the range of changes in object features caused by the noise. This ratio is calculated by comparing the changes caused by noise and the changes caused by the neural signals. The intensity of N3 represents the probability of failure each time an action is performed. Thus, it ranges from 0 to 1. The intensity of N4 represents the proportion of sensing error in the actual sensing result each time a perception occurs. Note that in the figure of Q, the lines of precision and specificity almost overlap.}
\label{fig:ablation}
\end{figure}

\subsubsection{\textbf{Results}}

We report the performance of our method ($p<0.05$, $p<0.01$, Bonferroni correction) and baseline ($\alpha=0.05, 0.01$), as shown in Table~\ref{tab:baseline_compare}.

(i) \textbf{\textit{Compared to baselines.}} The baseline method is more likely to be unconservative, which infers the property of Fisher randomization test \cite{Ding2023}: it is likely to identify more objects as the body than our method.
Our method achieves the best accuracy, precision, specificity, and F1 score in most tasks.  
Among our settings, the lower threshold leads to more strict standards, reaching better scores. 
When reducing the threshold of $p$, considerable increases are observed in accuracy, precision, specificity, and F1 score, while recall drops slightly. 
The Bonferroni correction brings an additional increase in accuracy and specificity in most tasks, as well as precision and F1 score (5 tasks increase).
However, it also causes an obvious drop in recall. 
The results of applying Bonferroni correction to the case where $\alpha=0.05$ and $\alpha=0.01$ are the same. It indicates that we are relatively close to the limit of correction.

(ii) \textbf{\textit{Parametric analysis.}}
Results (using T8 as the exemplary task) are illustrated in Figure~\ref{fig:ablation}.
As $Q$ increases, the performance first goes up and then drops. The peak appears when $Q$ is around 10. Note that the lines for precision and specificity overlap.
We see a slight decrease in performance when $N$ increases, implying that the number of objects has a minor influence on the performance.
Results also indicate that the performance gets better when $T$ increases, indicating that increased interaction opportunities lead to improved results. However, the performance growth rate slows down after $T>310$, indicating that there is a limit of $T$. 
Increases in the intensity of N1, N3, and N4 cause decreases in recall, precision, and F1, while accuracy and specificity remain stable.
The intensity of N4 represents the percentage error in observed data. It drops suddenly when intensity goes over 60\%. We can infer that our method can, to some extent, survive the partial view case, but global observation is still crucial to our method.
We do not observe an obvious drop in performance when increasing the intensity of other agent noise, indicating that our method is robust against such noise.

\subsection{Advanced Experiments: Mirror Test}

\begin{figure}[tb]
    \centering    
    \includegraphics[width=0.9\linewidth]{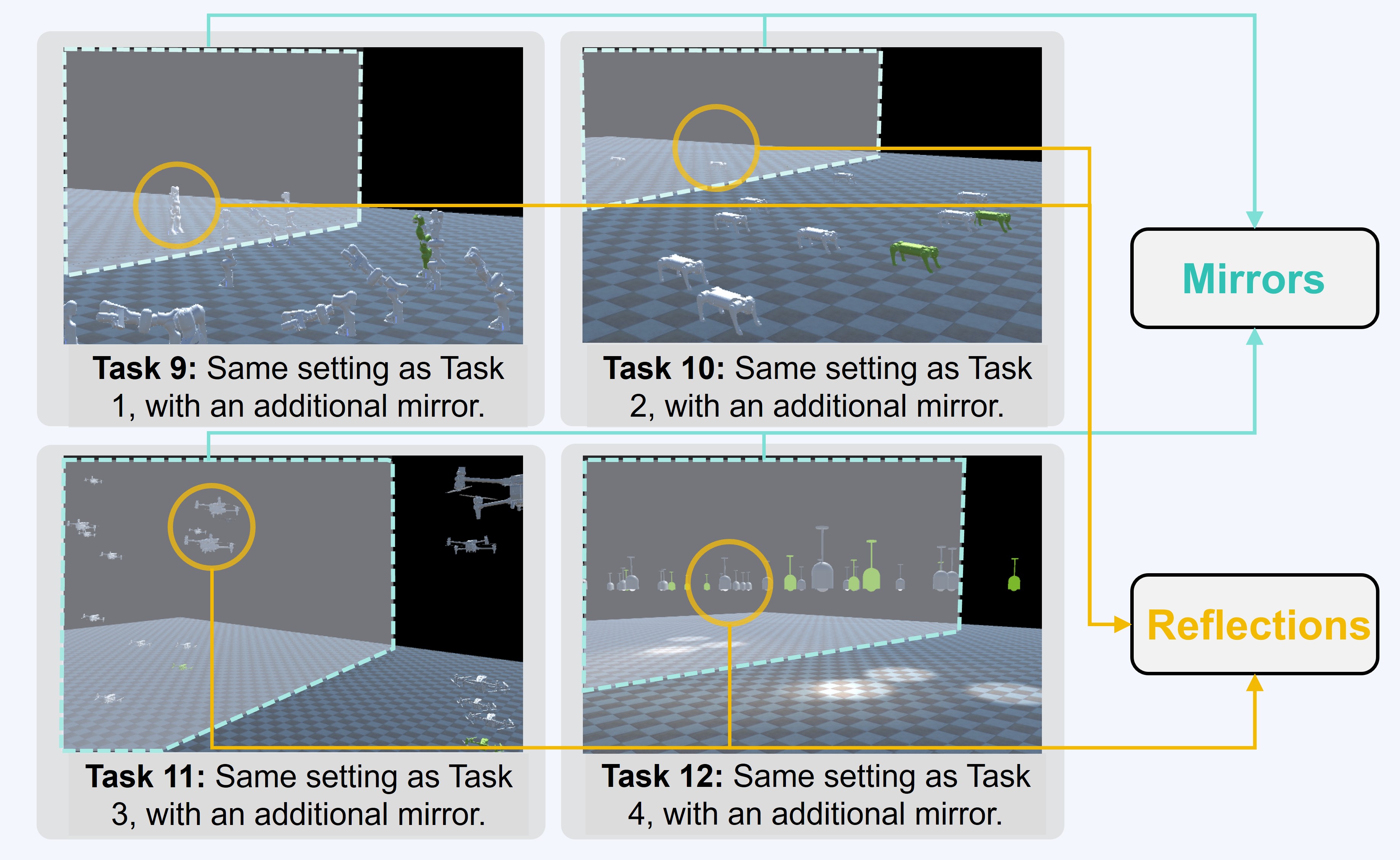}
    \caption{Simulated scenes for ``Mirror Test'', i.e., T9-T12. Their settings are similar to T1-T4, and the only difference is that each task has an additional mirror in the scene. }
    \label{fig:simulator_9_12}
\end{figure}

In the domain of cognitive science, the ``Mirror Test'' is a classic method to evaluate creatures' ability of self-cognition.
Only a few kinds of species can pass this test, such as humans and chimpanzees.
Therefore, in addition to the aforementioned 9 settings of general experimental scenarios, we further design four tasks (T9-T12) for the ``Mirror Test'', shown in Figure~\ref{fig:simulator_9_12}, following Gallup's settings~\cite{gallup1970chimpanzees}.

\subsubsection{\textbf{Experimental Settings}}
A mirror surface is generated in the room with a random location and a random horizontal direction, dividing the room into two parts.
The surface that this mirror lies on divides the room into two sub-rooms.
The half that faces the mirror is copied into the mirror world, with every object located in this area copied and denoted as reflections.
The other half, which is located at the back of the mirror, remains the same as originally designed.
The tasks of this test are to detect both the original bodies and the reflections.

\subsubsection{\textbf{Results}}
Results of the ``Mirror Test'' (T9-T12) are shown in Table~\ref{tab:mirror_test}. All tasks achieve good performance. For T10-T12, most metrics are over 90\%, indicating our method could handle the mirror task with features of 2d and 3d position, as well as the discrete state. Performance on T9 is slightly lower than others, indicating that the rotation feature is relatively difficult. Our method with Bonferroni correction shows the best performance on most metrics except recall. As for recall, using a threshold of 0.05 achieves the best score.
Drawing upon these tenets, we can, in part, adjudicate the capability of our method to pass the ``Mirror Test'' and take one step towards self-cognition.

\begin{table}[tb]
\setlength\tabcolsep{1.5pt}
\caption{Performance of our methods in the mirror tests.}
\label{tab:mirror_test}
\begin{center}

\begin{tabular}{c|c|ccccc}
\toprule
Task& Method & Accuracy & Recall & Precision & Specificity & F1 Score \\
\midrule
\multirow{3}{*}{T9} 
 & Ours ($p<0.05$) & 0.847 & \textbf{0.985} & 0.639 & 0.784 & 0.759 \\
 & Ours ($p<0.01$) & 0.866 & 0.977 & 0.699 & 0.812 & 0.797 \\
 & Ours (Bonferroni) & \textbf{0.871} & 0.973 & \textbf{0.709} & \textbf{0.819} & \textbf{0.805} \\
\midrule
\multirow{3}{*}{T10} 
 & Ours ($p<0.05$) & 0.907 & \textbf{0.971} & 0.594 & 0.896 & 0.711 \\
 & Ours ($p<0.01$) & 0.974 & 0.960 & 0.874 & 0.976 & 0.900 \\
 & Ours (Bonferroni) & \textbf{0.984} & 0.882 & \textbf{0.946} & \textbf{0.999} & \textbf{0.907} \\
\midrule
\multirow{3}{*}{T11} 
 & Ours ($p<0.05$) & 0.891 & \textbf{0.997} & 0.528 & 0.878 & 0.657 \\
 & Ours ($p<0.01$) & 0.979 & 0.987 & 0.863 & 0.978 & 0.905 \\
 & Ours (Bonferroni) & \textbf{0.995} & 0.948 & \textbf{0.970} & \textbf{1.000} & \textbf{0.956} \\
\midrule
\multirow{3}{*}{T12} 
 & Ours ($p<0.05$) & 0.942 & \textbf{1.000} & 0.652 & 0.936 & 0.753 \\
 & Ours ($p<0.01$) & 0.986 & \textbf{1.000} & 0.880 & 0.985 & 0.920 \\
 & Ours (Bonferroni) & \textbf{0.999} & 0.998 & \textbf{0.990} & \textbf{0.999} & \textbf{0.992} \\
\bottomrule
\end{tabular}
\end{center}
\end{table}

\section{Discussion}

In this work, we introduce a novel challenge named ``Body Discovery of Embodied AI'', and wish to tackle two fundamental dimensions of embodiment adaptation: (1) understanding the composition and boundaries of embodiments, and (2) grasping their functionality.
As a solution, we connect the causal inference method with body discovery tasks and present a causality-based approach. Then we compare the proposed approach with the naive method through simulations in various environments. The results indicate that our approach tackles the proposed challenge to a large extent.
Notably, our approach provides insights into the classic cognitive experiment of ``Mirror Test'', and help the embodied AI pass the test to some degree.

Next, we briefly review the related works about embodied cognition, embodied AI, and casual inference.

\subsection{Related Works}
\label{sec:related}

\textbf{Embodied Cognition.}
There is a movement afoot in cognitive science to grant the body a central role in shaping the mind \cite{wilson2002six,anderson2003embodied}. Early studies in cognition science focused on formal operations on abstract symbols whose connections to the outside world were of little
theoretical importance, like Fodor’s modularity hypothesis \cite{fodor1985precis}.
However, some researchers emphasized sensory and motor functions, as well as their
importance for successful interaction with the environment, like motor theories of perception \cite{liberman1985motor}, ecological psychology of Gibson \cite{cutting1982two}.  
Embodied cognition focuses on acting beings. People form cognition through interactions with environments~\cite{wilson2002six,anderson2003embodied}. 
The study on embodied cognition makes people recognize the necessity to rethink AI. 

\textbf{Embodied Artificial Intelligence.}
There is a rising trend of training intelligent agents with embodiments \cite{kaur2023review}. Researchers build virtual environments aiming at constructing AI-oriented simulators to accelerate the study of AI. For example, AI2-THOR~\cite{kolve2017ai2}, VirtualHome \cite{puig2018virtualhome}, Habitat \cite{savva2019habitat}, VRGym \cite{xie2019vrgym}, iGibson~\cite{xia2020interactive}, OmniGibson~\cite{li2023behavior}, and GRUtopia~\cite{wang2024grutopia} are popular virtual environments for assisting AI researches, which provides different kinds of embodiments that can be controlled by AI. Using these embodiments, AI agents can interact with the virtual environment, and thus learn various skills.

\textbf{Causal Inference.}
In intricate environments, estimating causal effects proves more effective than relying solely on associative relations across diverse domains, owing to their stability under varying conditions
\cite{imbens2015causal,Ding2023,pearl2009causality}. This approach provides robust evidence and mitigates data-related issues within a well-defined framework \cite{athey2017state}.
The gold standard in causal inference is the randomized experiment \cite{Morgan2012,Li2018}, traditionally analyzed through the potential outcomes framework. Within the potential outcomes framework \cite{Neyman1923,Rubin:1974}, four main methodologies exist: the Fisher randomization test, the Neyman approach, the Bayesian method, and the regression method \cite{imbens2015causal}. Numerous studies have explored the interplay between these methods, contributing to the literature on causal inference \cite{freedman2008regression_a,lin2013,LiDing2020}.

\subsection{Limitation and Future Work}
While the challenge framework employs a global observation paradigm, our proposed methodology retains functional compatibility with partial-view scenarios through minor sensor configuration adjustments. By implementing controlled stochastic sensor movements (e.g., integrating cameras with randomized rotational mechanisms on robotic platforms), the estimator maintains statistical unbiasedness while preserving operational efficacy. Nevertheless, substantial challenges persist in addressing generalized partial-view inference problems, particularly those involving nonlinear observation constraints, constituting valuable directions for subsequent research.

Complementing our statistical baseline evaluations, we empirically investigated the capacity of large language models (LLMs) for embodiment comprehension. When explicitly prompted with the body discovery challenge specifications, GPT-4~\cite{achiam2023gpt} demonstrated limited task engagement, typically responding with either architectural capability disclaimers or requests for explicit neural-embodiment mapping protocols. This behavioral pattern suggests fundamental limitations in current LLMs' ability to process embodied cognition tasks without explicit environmental grounding.

Given the inherent complexity of multi-agent system coordination, our current implementation remains confined to virtual simulation environments. This constraint underscores a critical research gap: while computational simulations provide necessary theoretical validation, physical environment deployment remains essential for authentic performance benchmarking. Future studies must therefore bridge this simulation-reality divide through systematic hardware-in-the-loop verification.

\bibliographystyle{abbrv}




\begin{thebibliography}{10}

\bibitem{achiam2023gpt}
J.~Achiam, S.~Adler, S.~Agarwal, L.~Ahmad, I.~Akkaya, F.~L. Aleman, D.~Almeida, J.~Altenschmidt, S.~Altman, S.~Anadkat, et~al.
\newblock Gpt-4 technical report.
\newblock {\em arXiv preprint arXiv:2303.08774}, 2023.

\bibitem{alaa2017review}
M.~Alaa, A.~A. Zaidan, B.~B. Zaidan, M.~Talal, and M.~L.~M. Kiah.
\newblock A review of smart home applications based on internet of things.
\newblock {\em Journal of Network and Computer Applications}, 97:48--65, 2017.

\bibitem{anderson2003embodied}
M.~L. Anderson.
\newblock Embodied cognition: A field guide.
\newblock {\em Artificial Intelligence}, 149(1):91--130, 2003.

\bibitem{athey2017state}
S.~Athey and G.~W. Imbens.
\newblock The state of applied econometrics: Causality and policy evaluation.
\newblock {\em Journal of Economic perspectives}, 31(2):3--32, 2017.

\bibitem{baum2017survey}
S.~Baum.
\newblock A survey of artificial general intelligence projects for ethics, risk, and policy.
\newblock {\em Global Catastrophic Risk Institute Working Paper}, pages 17--1, 2017.

\bibitem{cully2015robots}
A.~Cully, J.~Clune, D.~Tarapore, and J.-B. Mouret.
\newblock Robots that can adapt like animals.
\newblock {\em Nature}, 521(7553):503--507, 2015.

\bibitem{cutting1982two}
J.~E. Cutting.
\newblock Two ecological perspectives: Gibson vs. shaw and turvey.
\newblock {\em The American Journal of Psychology}, pages 199--222, 1982.

\bibitem{diaz2023machine}
F.~D{\'\i}az~Ledezma and S.~Haddadin.
\newblock Machine learning--driven self-discovery of the robot body morphology.
\newblock {\em Science Robotics}, 8(85):eadh0972, 2023.

\bibitem{Ding2023}
P.~Ding.
\newblock A first course in causal inference.
\newblock {\em arXiv preprint arXiv:2305.18793}, 2023.

\bibitem{ding2024first}
P.~Ding.
\newblock {\em A first course in causal inference}.
\newblock CRC Press, 2024.

\bibitem{driess2023palm}
D.~Driess, F.~Xia, M.~S. Sajjadi, C.~Lynch, A.~Chowdhery, B.~Ichter, A.~Wahid, J.~Tompson, Q.~Vuong, T.~Yu, et~al.
\newblock Palm-e: An embodied multimodal language model.
\newblock {\em arXiv preprint arXiv:2303.03378}, 2023.

\bibitem{duan2022survey}
J.~Duan, S.~Yu, H.~L. Tan, H.~Zhu, and C.~Tan.
\newblock A survey of embodied ai: From simulators to research tasks.
\newblock {\em IEEE Transactions on Emerging Topics in Computational Intelligence}, 6(2):230--244, 2022.

\bibitem{fodor1985precis}
J.~A. Fodor.
\newblock Precis of the modularity of mind.
\newblock {\em Behavioral and Brain Sciences}, 8(1):1--5, 1985.

\bibitem{freedman2008regression_a}
D.~A. Freedman.
\newblock On regression adjustments in experiments with several treatments.
\newblock {\em The Annals of Applied Statistics}, 2:176--196, 2008.

\bibitem{gallup1970chimpanzees}
G.~G. Gallup~Jr.
\newblock Chimpanzees: self-recognition.
\newblock {\em Science}, 167(3914):86--87, 1970.

\bibitem{gallup2002mirror}
G.~G. Gallup~Jr, J.~R. Anderson, and D.~J. Shillito.
\newblock The mirror test.
\newblock {\em The cognitive animal: Empirical and theoretical perspectives on animal cognition}, pages 325--333, 2002.

\bibitem{goertzel2014artificial}
B.~Goertzel.
\newblock Artificial general intelligence: concept, state of the art, and future prospects.
\newblock {\em Journal of Artificial General Intelligence}, 5(1):1, 2014.

\bibitem{goertzel2007artificial}
B.~Goertzel and C.~Pennachin.
\newblock {\em Artificial general intelligence}, volume~2.
\newblock Springer, 2007.

\bibitem{gupta2021embodied}
A.~Gupta, S.~Savarese, S.~Ganguli, and L.~Fei-Fei.
\newblock Embodied intelligence via learning and evolution.
\newblock {\em Nature Communications}, 12(1):5721, 2021.

\bibitem{imbens2015causal}
G.~W. Imbens and D.~B. Rubin.
\newblock {\em Causal {I}nference for {S}tatistics, {S}ocial, and {B}iomedical {S}ciences: {A}n {I}ntroduction}.
\newblock New York: Cambridge University Press, 2015.

\bibitem{kaur2023review}
D.~P. Kaur, N.~P. Singh, and B.~Banerjee.
\newblock A review of platforms for simulating embodied agents in 3d virtual environments.
\newblock {\em Artificial Intelligence Review}, 56(4):3711--3753, 2023.

\bibitem{kiverstein2012meaning}
J.~Kiverstein.
\newblock The meaning of embodiment.
\newblock {\em Topics in Cognitive Science}, 4(4):740--758, 2012.

\bibitem{kolve2017ai2}
E.~Kolve, R.~Mottaghi, D.~Gordon, Y.~Zhu, A.~Gupta, and A.~Farhadi.
\newblock Ai2-thor: An interactive 3d environment for visual ai.
\newblock {\em arXiv preprint arXiv:1712.05474}, 2017.

\bibitem{lee2006physically}
K.~M. Lee, Y.~Jung, J.~Kim, and S.~R. Kim.
\newblock Are physically embodied social agents better than disembodied social agents?: The effects of physical embodiment, tactile interaction, and people's loneliness in human--robot interaction.
\newblock {\em International Journal of Human-Computer Studies}, 64(10):962--973, 2006.

\bibitem{li2023behavior}
C.~Li, R.~Zhang, J.~Wong, C.~Gokmen, S.~Srivastava, R.~Mart{\'\i}n-Mart{\'\i}n, C.~Wang, G.~Levine, M.~Lingelbach, J.~Sun, et~al.
\newblock Behavior-1k: A benchmark for embodied ai with 1,000 everyday activities and realistic simulation.
\newblock In {\em Proceedings of the Conference on Robot Learning}, pages 80--93. PMLR, 2023.

\bibitem{LiDing2020}
X.~Li and P.~Ding.
\newblock Rerandomization and regression adjustment.
\newblock {\em Journal of the Royal Statistical Society, Series B}, 82:241--268, 2020.

\bibitem{Li2018}
X.~Li, P.~Ding, and D.~B. Rubin.
\newblock Asymptotic theory of rerandomization in treatment-control experiments.
\newblock volume 115, pages 9157--9162, Sep 11 2018.

\bibitem{liberman1985motor}
A.~M. Liberman and I.~G. Mattingly.
\newblock The motor theory of speech perception revised.
\newblock {\em Cognition}, 21(1):1--36, 1985.

\bibitem{lin2013}
W.~Lin.
\newblock {Agnostic notes on regression adjustments to experimental data: Reexamining Freedman's critique}.
\newblock {\em The Annals of Applied Statistics}, 7:295--318, 2013.

\bibitem{minsky1988society}
M.~Minsky.
\newblock {\em Society of mind}.
\newblock Simon and Schuster, 1988.

\bibitem{Morgan2012}
K.~L. Morgan and D.~B. Rubin.
\newblock Rerandomization to improve covariate balance in experiments.
\newblock {\em The Annals of Statistics}, 40:1263--1282, 2012.

\bibitem{pearl2009causality}
J.~Pearl.
\newblock {\em Causality}.
\newblock Cambridge university press, 2009.

\bibitem{pei2019towards}
J.~Pei, L.~Deng, S.~Song, M.~Zhao, Y.~Zhang, S.~Wu, G.~Wang, Z.~Zou, Z.~Wu, W.~He, et~al.
\newblock Towards artificial general intelligence with hybrid tianjic chip architecture.
\newblock {\em Nature}, 572(7767):106--111, 2019.

\bibitem{peng2024tong}
Y.~Peng, J.~Han, Z.~Zhang, L.~Fan, T.~Liu, S.~Qi, X.~Feng, Y.~Ma, Y.~Wang, and S.-C. Zhu.
\newblock The tong test: Evaluating artificial general intelligence through dynamic embodied physical and social interactions.
\newblock {\em Engineering}, 34:12--22, 2024.

\bibitem{pfeifer2006body}
R.~Pfeifer and J.~Bongard.
\newblock {\em How the body shapes the way we think: a new view of intelligence}.
\newblock MIT press, 2006.

\bibitem{puig2018virtualhome}
X.~Puig, K.~Ra, M.~Boben, J.~Li, T.~Wang, S.~Fidler, and A.~Torralba.
\newblock Virtualhome: Simulating household activities via programs.
\newblock In {\em Proceedings of the IEEE Conference on Computer Vision and Pattern Recognition (CVPR)}, pages 8494--8502, 2018.

\bibitem{rajan2017towards}
K.~Rajan and A.~Saffiotti.
\newblock Towards a science of integrated ai and robotics, 2017.

\bibitem{Rubin:1974}
D.~B. Rubin.
\newblock Estimating causal effects of treatments in randomized and nonrandomized studies.
\newblock {\em Journal of Educational Psychology}, 66:688--701, 1974.

\bibitem{saegusa2010self}
R.~Saegusa, G.~Metta, and G.~Sandini.
\newblock Self-body discovery based on visuomotor coherence.
\newblock In {\em Proceedings of the 3rd International Conference on Human System Interaction}, pages 356--362. IEEE, 2010.

\bibitem{savva2019habitat}
M.~Savva, A.~Kadian, O.~Maksymets, Y.~Zhao, E.~Wijmans, B.~Jain, J.~Straub, J.~Liu, V.~Koltun, J.~Malik, et~al.
\newblock Habitat: A platform for embodied ai research.
\newblock In {\em Proceedings of International Conference on Computer Vision (ICCV)}, pages 9339--9347, 2019.

\bibitem{shah2021soft}
D.~S. Shah, J.~P. Powers, L.~G. Tilton, S.~Kriegman, J.~Bongard, and R.~Kramer-Bottiglio.
\newblock A soft robot that adapts to environments through shape change.
\newblock {\em Nature Machine Intelligence}, 3(1):51--59, 2021.

\bibitem{Neyman1923}
J.~Splawa-Neyman.
\newblock On the application of probability theory to agricultural experiments. \textsc{E}ssay on principles (with discussion). \textsc{S}ection 9. \textsc{R}eprinted.
\newblock {\em Statistical Science}, 5:465--472, 1923.

\bibitem{Neyman:1923}
J.~Splawa-Neyman, D.~M. Dabrowska, and T.~P. Speed.
\newblock On the application of probability theory to agricultural experiments. essay on principles. section 9.
\newblock {\em Statistical Science}, 5:465--472, 1990.

\bibitem{sun2023neighbor}
Z.~Sun, Q.~Liang, M.~Wang, and Z.~Zhang.
\newblock Neighbor-environment observer: An intelligent agent for immersive working companionship.
\newblock In {\em Proceedings of the 36th Annual ACM Symposium on User Interface Software and Technology}, pages 1--14, 2023.

\bibitem{vemprala2023chatgpt}
S.~Vemprala, R.~Bonatti, A.~Bucker, and A.~Kapoor.
\newblock Chatgpt for robotics: Design principles and model abilities.
\newblock {\em Microsoft Auton. Syst. Robot. Res}, 2:20, 2023.

\bibitem{wang2024grutopia}
H.~Wang, J.~Chen, W.~Huang, Q.~Ben, T.~Wang, B.~Mi, T.~Huang, S.~Zhao, Y.~Chen, S.~Yang, et~al.
\newblock Grutopia: Dream general robots in a city at scale.
\newblock {\em arXiv preprint arXiv:2407.10943}, 2024.

\bibitem{wilson2002six}
M.~Wilson.
\newblock Six views of embodied cognition.
\newblock {\em Psychonomic Bulletin \& Review}, 9:625--636, 2002.

\bibitem{xia2020interactive}
F.~Xia, W.~B. Shen, C.~Li, P.~Kasimbeg, M.~E. Tchapmi, A.~Toshev, R.~Mart{\'\i}n-Mart{\'\i}n, and S.~Savarese.
\newblock Interactive gibson benchmark: A benchmark for interactive navigation in cluttered environments.
\newblock {\em IEEE Robotics and Automation Letters}, 5(2):713--720, 2020.

\bibitem{xie2019vrgym}
X.~Xie, H.~Liu, Z.~Zhang, Y.~Qiu, F.~Gao, S.~Qi, Y.~Zhu, and S.-C. Zhu.
\newblock Vrgym: A virtual testbed for physical and interactive ai.
\newblock In {\em Proceedings of the ACM Turing Celebration Conference-China}, pages 1--6, 2019.

\bibitem{yu2008morpho}
C.-H. Yu, K.~Haller, D.~Ingber, and R.~Nagpal.
\newblock Morpho: A self-deformable modular robot inspired by cellular structure.
\newblock In {\em Proceedings of the IEEE/RSJ International Conference on Intelligent Robots and Systems}, pages 3571--3578. IEEE, 2008.

\bibitem{zhang2019symmetrical}
Z.~Zhang, C.~Wang, D.~Weng, Y.~Liu, and Y.~Wang.
\newblock Symmetrical reality: Toward a unified framework for physical and virtual reality.
\newblock In {\em Proceedings of the IEEE Conference on Virtual Reality and 3D User Interfaces (VR)}, pages 1275--1276. IEEE, 2019.

\bibitem{zhang2024emergence}
Z.~Zhang, Z.~Zhang, Z.~Jiao, Y.~Su, H.~Liu, W.~Wang, and S.-C. Zhu.
\newblock On the emergence of symmetrical reality.
\newblock In {\em Proceedings of the IEEE Conference Virtual Reality and 3D User Interfaces (VR)}, pages 639--649. IEEE, 2024.

\end{thebibliography}

\end{document}